\documentclass[letterpaper, 10 pt, conference]{ieeeconf}  

\IEEEoverridecommandlockouts                              

\usepackage{graphicx} 
\usepackage{subcaption}
\usepackage{kotex}
\usepackage{float}
\usepackage{multirow}
\usepackage{adjustbox}
\usepackage{algorithmic}
\usepackage{algorithm}
\usepackage{amsmath} 
\usepackage{amssymb}  

\usepackage[font=footnotesize,labelfont=footnotesize]{subcaption}
\usepackage[font=footnotesize,labelfont=footnotesize]{caption}
\usepackage{xcolor}
\usepackage{comment}
\usepackage{xspace}

\definecolor{darkblue}{rgb}{0.15,0.15,0.55}
\definecolor{lightgrey}{rgb}{0.75,0.75,0.75}

\providecommand{\codecomment}[1]{\textcolor{lightgrey}{\dotfill//\,}\textcolor{darkblue}{\textrm{#1}}}

\DeclareMathOperator*{\argmin}{arg\,min}
\DeclareMathOperator*{\argmax}{arg\,max}

\newcommand{\nphard}{$\mathcal{NP}$-hard\xspace}

\newcommand{\thm}{\noindent \textbf{Theorem}\xspace}

\newcommand{\pf}{\noindent \textbf{Proof. }\xspace}

\newcommand{\qedm}{\hfill $\square$}

\title{\LARGE \bf Where to relocate?: Object rearrangement inside cluttered\\and confined environments for robotic manipulation}

\author{Sang Hun Cheong, Brian Y. Cho, Jinhwi Lee, ChangHwan Kim$^*$, and Changjoo Nam$^*$
\thanks{This work was supported by the Technology Innovation Program and Industrial Strategic Technology Development Program (10077538, Development of manipulation technologies in social contexts for human-care service robots). The authors are with Korea Institute of Science and Technology.
$^*$Corresponding authors: \texttt{ \{ckim, cjnam\}@kist.re.kr}. }}

\begin{document}

\maketitle
\thispagestyle{empty}
\pagestyle{empty}

\begin{abstract}

We present an algorithm determining \textit{where to relocate} objects inside a cluttered and confined space while rearranging objects to retrieve a target object. Although methods that decide \textit{what to remove} have been proposed, planning for the placement of removed objects inside a workspace has not received much attention. Rather, removed objects are often placed outside the workspace, which incurs additional laborious work (e.g., motion planning and execution of the manipulator and the mobile base, perception of other areas). Some other methods manipulate objects only inside the workspace but without a principle so the rearrangement becomes inefficient.

In this work, we consider both monotone (each object is moved only once) and non-monotone arrangement problems which have shown to be \nphard. Once the sequence of objects to be relocated is given by any existing algorithm, our method aims to minimize the number of pick-and-place actions to place the objects until the target becomes accessible. From extensive experiments, we show that our method reduces the number of pick-and-place actions and the total execution time (the reduction is up to 23.1\% and 28.1\% respectively) compared to baseline methods while achieving higher success rates.

\end{abstract}


\section{Introduction}

Robots tasked with retrieving objects from cluttered and confined environments need to accomplish several processes: perception in the presence of occlusion, task planning for object relocation, motion planning, grasp planning, and etc. Among these, relocation planning determines what to remove to where in order to retrieve the target object without collisions. In spite of much attention on relocation planning, most work has focused on determining the objects to be relocated (and their orders) to secure a collision-free path of the robot manipulator~\cite{han2018complexity,lee2019efficient,krontiris2015dealing}. On the other hand, the placement of the relocated objects is often done without a principle or predetermined via hard-coding. 

\begin{figure}[t!]
    \centering
    \includegraphics[scale=0.038]{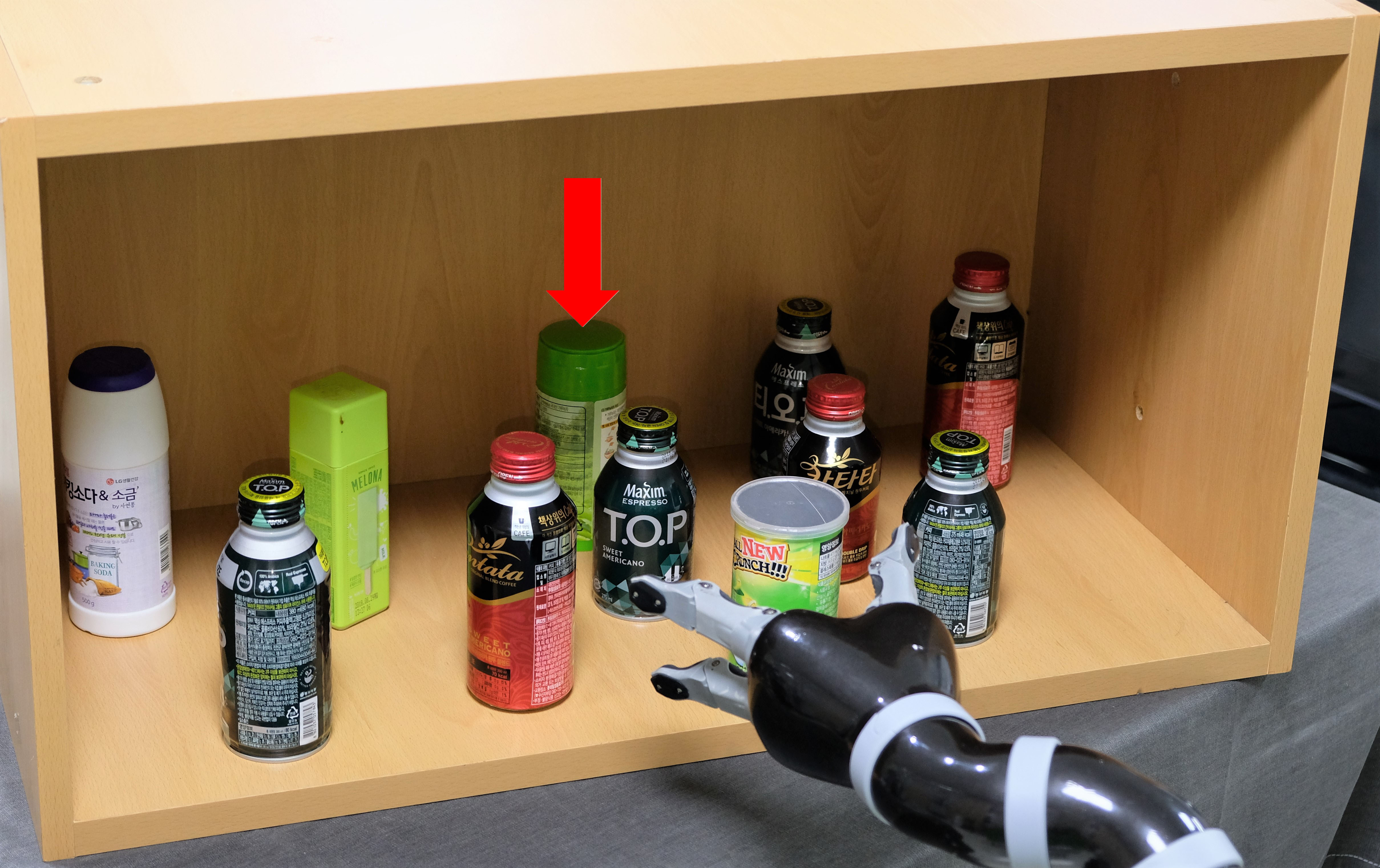}
    \caption{A cluttered and confined workspace where the manipulator try to retrieve the target object (red arrow)}
    \label{fig:clutter}\vspace{-13pt}
\end{figure}

As also pointed out in \cite{han2018complexity}, manipulating objects only inside the workspace which is cluttered and confined (as shown in Fig.~\ref{fig:clutter}) can enhance efficiency of target retrieval by saving the effort to access buffer spaces outside the workspace (e.g., the planning and execution time to move the manipulator and/or the mobile base, the time for perceiving the buffer spaces). Moreover, robots may not be able to find buffer spaces nearby the workspace. 

In this work, we consider the problem determining the placement of relocated objects inside a cluttered and confined workspace. The problem considers not \textit{what to relocate} but \textit{where to place}. As a result, the robot can retrieve the target object from clutter without reserving some buffer spaces outside the current workspace occupied by objects. We assume that the sequence of objects to be relocated (i.e., what to relocate in what order) is computed upfront using one of existing methods like~\cite{lee2019efficient,nam2020fast,nam2019planning}. Based on the sequence, we aim to plan for rearrangement to place all objects in the sequence inside the workspace with the minimum number of pick-and-place actions. We consider the kinematic constraints of the manipulator so the result should be executable by real robots. Also, we consider both monotone and non-monotone object manipulation problems (i.e., each object is moved only once and some objects need multiple moves, respectively), which are known to be \nphard~\cite{stilman2007manipulation}. Dealing with the non-monotone problem is important since the robot can utilize empty spaces, which are scarce in our setting, as much as possible. 


Some previous work considers less congested environments so placing objects to arbitrary empty spaces does not interrupt the robot with performing the relocation plan~\cite{van2009path,dogar2012planning,lin2015planning,vega2016asymptotically}. Some other work dealing with dense clutter does not solve the problem of where to place but use some predetermined goal poses of objects~\cite{krontiris2015dealing}. Other works~\cite{garrett2015backward,haustein2015kinodynamic,moll2018randomized} also focus on determining what to remove in what order only while not considering where to place the removed objects. In~\cite{han2018efficient}, objects are stored in stacks so the robot should decides the stack to place an object. However, the environments considered in our work has a structural difference so the last-in first-out principle cannot be used. The tabletop arrangement problem in~\cite{han2018complexity} considers buffer locations to place objects but they are assumed to be located outside the tabletop. In~\cite{havur2014geometric}, a final configuration for rearrangement is determined without considering the reachability of the spaces and the robot kinematics so would not be achieved when working with real robots. 

Our method aims to find a plan for rearrangement quickly while only using spaces inside a cluttered workspace occupied by objects. Initially, empty spaces in the workspace are converted into discrete buffer slots where objects can be placed. Then the method decides if the given instance falls into the solvable class depending on the numbers of objects to be relocated and the buffer slots. Then, we introduce a principle to choose the slot to be used for each relocated object. The choice of slots is done while considering the kinematic constraints of the robot so a solution executable by the robot is computed.  

The following are the contribution of this work: 
\begin{itemize}
  \item We propose a polynomial-time algorithm to decide where to place objects that are removed to generate a collision-free path to a target object. Our method uses the space only inside a cluttered and confined workspace. 
  \item Our method takes the task and motion planning (TAMP) approach~\cite{dantam2018task} so task plans in the discrete space (i.e., rearranging objects among buffer slots) incorporate motions in the continuous space. Thus, the plans should be executable by robots.
  \item We provide extensive experiments in simulated environments and comparisons with baseline methods showing that our method can reduce the number of pick-and-place actions and total execution time.
\end{itemize}
\vspace{-5pt}

\section{Problem Description}
\label{sec:prob}
We consider a workspace $\mathcal{W}$ occupied by a set of $N$ objects $\mathcal{O} = \{o_1, \cdots, o_N\}$ and a target object $o_t \notin \mathcal{O}$. We assume that $\mathcal{W}$ is bounded by immovable obstacles (like shelves and walls) so defined by its size $l_w \times l_d \times l_h$. As the workspace is confined, the robot can approach the objects from only one side (so overhand grasps are not allowed). In this environment, the objects in $\mathcal{O}_R \subseteq \mathcal{O}$ should be relocated to secure a collision-free path to a target object from the end-effector. We assume that such the objects $\mathcal{O}_R$ are precomputed using any existing algorithm like~\cite{lee2019efficient,nam2019planning}. 

We aim to minimize the number of pick-and-place actions $k$ while rearranging the objects in $\mathcal{O}_R$ only inside $\mathcal{W}$. The objective value is directly related to the total execution time until completing the retrieval task because inefficient rearrangement may bring unnecessary moves of objects. We assume that the geometry of the objects (e.g., size and pose) is known and does not change while the robot is performing an action. We also assume that objects do not have restrictions in their grasping direction so model objects using cylinders. No pushing or dragging action is used owing to the difficulty of planning and executing such actions in cluttered and confined environments.

We consider both monotone and non-monotone problems. The non-monotone problem is considered more difficult because the number of cases branching from each action is explosive as rolling back some previous actions is allowed. Even though the monotone problem is known to be easier, it is shown to be \nphard~\cite{stilman2007manipulation}. Thus, our goal is to develop an efficient method that solves the both problems in polynomial time. In addition, the resulting relocation plan should be executable by a robot manipulator so including motion planning is indispensable for practical uses. 

\section{The rearrangement planning algorithm}
\label{sec:alg}
In this section, we give an overview of the proposed rearrangement planning algorithm (the pseudocode is shown in Alg.~\ref{alg:rearrange}) followed by detailed descriptions. We also provide example executions of the monotone and non-monotone problems.

In the beginning, the rearrangement planning algorithm finds empty buffer slots (\textit{candidate slots} denoted by $\mathcal{S}_c$) that objects could fit into. They are found without considering the kinematic constraints of the manipulator so not all candidate slots are reachable. Thus, motion planning follows to obtain a list of reachable candidate slots (\textit{valid candidate slots} denoted by $\mathcal{S}_v$). For $s \in \mathcal{S}_v$, we introduce a quantity $\beta(s) \in \mathbb{Z}_{\ge 0}$ representing the number of valid slots that could be occluded if an object is placed in $s$. The placement of objects in $\mathcal{O}_R$ are determined by choosing $s_i$ such that $s_i = \argmin_{s_i \in \mathcal{S}_v} \beta(s_i)$. This principle increases the efficiency of rearrangement as the number of remaining valid candidate slots is maximized. The algorithm is shown to run in polynomial time.

\subsection{Algorithm descriptions}

\subsubsection{Preliminaries} 
\label{sec:pre}
We employ two existing algorithms. The first one is the modified VFH+~\cite{lee2019efficient} that is used to check if an object (or a candidate slot) in clutter is accessible by the robot end-effector. The output of the algorithm shows if an object grasped by the end-effector does not collide with other objects. Since the collision check does not consider the whole robot arm, a motion planner would be more appropriate to check if the whole arm does not collide. However, motion planning for the pick-and-place action for every object or slot requires significant computation time. Thus, we use the modified VFH+ (running in $O(N^2)$) to filter out inaccessible objects/slots quickly to save the motion planning time. 

The second algorithm employed is a task planning method determining the sequence of objects to be relocated~\cite{nam2019planning}. It runs in $O(N^4)$ and is used once before our method runs to obtain $\mathcal{O}_R$. During relocation, the algorithm could be used several times if a problem instance falls into the non-monotone class.  


\subsubsection{Finding candidate slots $\mathcal{S}_c$}
\label{sec:findc}
\begin{figure}[h!]
\vspace{-5pt}
    \captionsetup{skip=0pt}
    \centering
    \begin{subfigure}{0.217\textwidth}
    \captionsetup{skip=0pt}
    \centering
	    \includegraphics[width=0.6\textwidth]{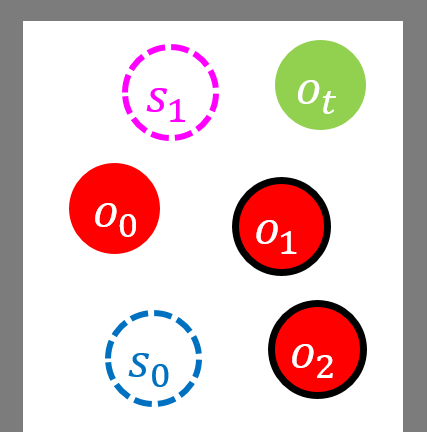}
        \caption{$s_0$ is valid as it is reachable and does not occlude the objects in $\mathcal{O}_R = \{o_2, o_1\}$. The back one $s_1$ is not reachable so invalid.}
        \label{fig:candidate1}
    \end{subfigure}\qquad
    \begin{subfigure}{0.228\textwidth}
    \captionsetup{skip=0pt}
    \centering
	    \includegraphics[width=0.571\textwidth]{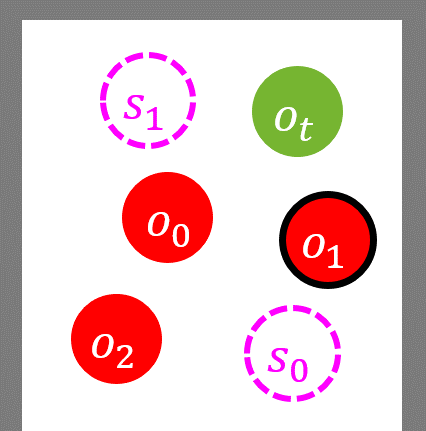}
        \caption{$s_0$ is not valid since it occludes the object in $\mathcal{O}_R = \{o_1\}$ so prevents clearing obstacles to grasp $o_t$. The back one $s_1$ is not reachable so invalid.}
        \label{fig:candidate2}
    \end{subfigure}
    \caption{Examples of candidate slots $s_0$ and $s_1$. The magenta dotted outline represents invalid slots while the blue means valid slots. The bold outline represents $\mathcal{O}_R$ that are objects to be relocated.}
  \label{fig:candidate}
\vspace{-10pt}
\end{figure}

First, we need to find empty spaces in $\mathcal{W}$ where objects can be placed during relocation ($s_0$ and $s_1$ in Fig.~\ref{fig:candidate1}). We generate discrete empty slots, which we call candidate slots, from the empty spaces. The set of slots $\mathcal{S}_c$ where $|\mathcal{S}_v| = M$ can be found by various methods. As deciding if a set of circles can be packed into a square is \nphard~\cite{demaine2010circle}, there have been many efficient approximation algorithms as reviewed in \cite{hifi2009literature}. Since the focus of this work is not finding the slots, we use a simple sampling-based approach that can be implemented conveniently. Specifically, a circle (which can contain any object in the workspace) is sampled from the empty space in $\mathcal{W}$.\footnote{We assume that the largest object determining the size of the slots is moderate. Objects that are larger than what the robot can grasp are considered immovable.} Once a circle is sampled, the area occupied by the circle is considered unempty. Sampling of circles is terminated after 1000 trials which takes less than 0.5 sec. The number of trials for sampling can be adjusted depending on the time allowed. After finding candidate slots, the algorithm decides if a problem instance is solvable. If $|\mathcal{S}_c| < |\mathcal{O}_R|$, the instance cannot be solved because candidate slots are insufficient to place all objects in $\mathcal{O}_R$ (lines~2--4 in Alg.~\ref{alg:rearrange}).

\subsubsection{Finding valid candidate slots $\mathcal{S}_v$}
\label{sec:findvc}
Not all slots in $\mathcal{S}_c$ are reachable by the robot since some objects may occlude them ($s_1$ in Fig.~\ref{fig:candidate1}). Thus, they should be checked if the robot (end-effector as well as the arm) can reach the slots without collisions. A straightforward way to check reachability is planning for motions of the manipulator. If a feasible motion is found, it means that the slot is reachable. However, planning motions for all $M$ candidate slots would take long so we screen out some of them by checking if only the end-effector can reach the slots. The screening is done using the modified VFH+ which often runs faster than motion planning of the whole arm. In addition, the slots that occlude any object in $\mathcal{O}_R$ are not valid since doing so will prevent clearing the path to the target ($s_0$ in Fig.~\ref{fig:candidate2}).

\subsubsection{Computing $\beta(s)$}
\label{sec:beta}
To rearrange all objects in $\mathcal{O}_R$, placing objects to valid candidate slots should be done carefully. Otherwise, a misplace of an object would reduce valid slots that are supposed to be used in the succeeding rearrangements. We introduce a quantity $\beta(s)$ for $s \in \mathcal{S}_v$ that represents the number of valid slots becoming invalid if an object is placed to $s$. If an object is relocated to a slot whose $\beta$ is the minimum, the number of remaining valid slots is maximized. Thus, the slots are utilized efficiently. Technically, $\beta(s)$ computation is done using the modified VFH+ to examine the number of slots that are invalidated if an object is assumed to be placed in $s$.

\subsubsection{Full algorithm}
\label{sec:overall}
Overall, the principle for rearrangement is simple as follows. Every step, the algorithm relocates an object to valid slots whose $\beta$ is the minimum. If the number of valid slots is not enough to rearrange all objects in $\mathcal{O}_R$, the algorithm secures additional valid candidate slots. It is done by moving some objects to invalid slots. Then the spaces that were occupied by the moved objects becomes valid slots in return. This additional rearrangement to secure valid slots is also done by following the principle based on $\beta$.

In Alg.~\ref{alg:rearrange}, $\mathcal{O}_R$ is computed (line~1) using the method proposed in~\cite{nam2019planning}. Then the algorithm checks if the problem instance is solvable. If $|\mathcal{S}_c|$ is less than $|\mathcal{O}_R|$, the algorithm terminates since the workspace has insufficient empty space for rearranging objects (lines~2--4).\footnote{One may increase $|\mathcal{S}_c|$ by running more trials in the process described in Sec.~\ref{sec:findc}.} 

\begin{figure*}[t]
    \captionsetup{skip=0pt}
    \centering
    \begin{subfigure}{0.95\textwidth}
    \captionsetup{skip=0pt}
    \centering
	    \includegraphics[width=0.65\textwidth]{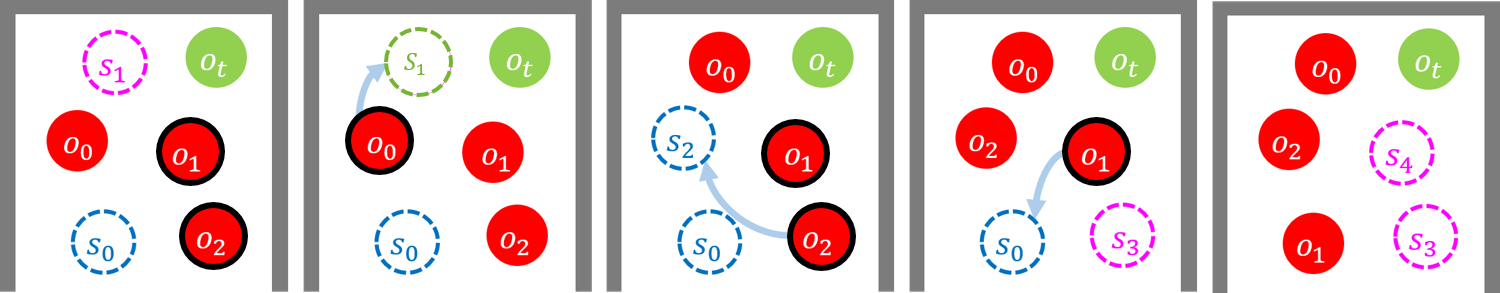}
        \caption{An example execution of Alg.~\ref{alg:rearrange} for an instance which belongs to the monotone class. In the beginning, valid ($s_0$) and invalid ($s_1$) candidate slots are found. Since the number of valid slots is insufficient to place $\mathcal{O}_R = \{o_2, o_1\}$, an additional valid slot should be obtained. It is done by moving $o_0$ to $s_1$. Then $o_2$ is moved to $s_2$ since $\beta(s_2) = 0$ is the minimum ($\beta(s_0) = 1$). Finally, $o_1$ is moved to $s_0$ because $s_3$ is invalid as it occludes the path to $o_t$.}
        \label{fig:monotone_example}
    \end{subfigure}\\
    \begin{subfigure}{0.95\textwidth}
    \captionsetup{skip=0pt}
    \centering
	    \includegraphics[width=0.73\textwidth]{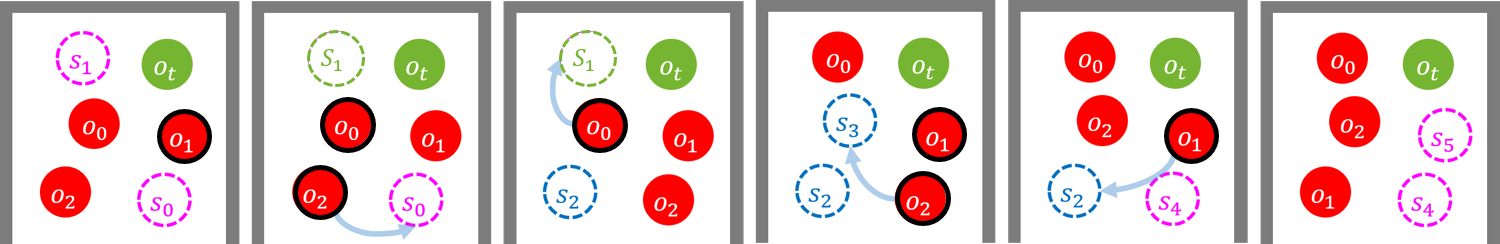}
        \caption{An example execution of Alg.~\ref{alg:rearrange} for an instance which belongs to the non-monotone class. In the beginning, there is no valid slots since $s_1$ is not reachable and $s_0$ occludes the path to the target and $\mathcal{O}_R$. However, the algorithm continues since $|\mathcal{O}_R| \le |\mathcal{S}_c|$. The algorithm decides to obtain a new valid slot by moving an object to $s_1$ which becomes $s^\prime$. The objects to be relocated to use $s^\prime$ is $\mathcal{O}^\prime_R = \{o_2, o_0\}$.  $o_2$ is moved to $s_0$ as it is a valid slot for the new relocation plan $\mathcal{O}^\prime_R$ ($\beta^\prime(s_0) = 0$). Then $o_0$ is moved to $s_1$ so valid slots $s_2$ and $s_3$ are obtained. The rest is the same with Fig.~\ref{fig:monotone_example} which simply follows the principle using $\beta$.}
        \label{fig:non_monotone_example}
    \end{subfigure}
    \caption{Example executions of Alg.~\ref{alg:rearrange}. Red and green circles represent obstacles and the target, respectively. The dotted circles represent candidate slots where the magenta, blue and green outlines indicate invalid, valid slots and the invalid slot with the minimum number of occluding objects, respectively. The bold outline represents objects to be relocated.}
  \label{fig:example_execution}
\vspace{-10pt}
\end{figure*}

In each iteration of the loop (lines~5--33), valid candidate slots $\mathcal{S}_v$ are updated to reflect the rearrangement done in the previous iteration (line~6). If the number of valid slots is smaller than the number of objects to be relocated, the robot needs to acquire additional valid slots. It can be done by placing objects into invalid candidate slots that are currently occluded. In Figs.~\ref{fig:monotone_example} and \ref{fig:non_monotone_example}, $s_1$ is such an invalid slot. By placing an object to $s_1$, a valid slot $s_2$ is obtained in both Figs.~\ref{fig:monotone_example} and \ref{fig:non_monotone_example}. If the set of such invalid slots $\mathcal{S}_e$ is empty, it means that there is no chance to generate additional valid slots. In this case, the algorithm terminates (line~10). 

Then rearranging objects occluding one of the slot in $\mathcal{S}_e$ follows to obtain a new valid slot. In order to minimize the number of pick-and-place actions to obtain a new valid slot, we choose $s^\prime$ that is the invalid slot with the minimum number of occluding objects. The choice of $s^\prime$ is done by calling \textsc{RelocatePlan} for all $s_i \in \mathcal{S}_e$ and comparing $|\mathcal{O}^i_R|$ (line~12 encapsulates the loop with an $\argmax$ operation for simplicity). The new sequence $\mathcal{O}^\prime_R$ in line~13 has objects to be relocated to access $s^\prime$.
If only one object occludes $s^\prime$ (like $s_1$ in Fig.~\ref{fig:monotone_example}), the problem instance falls into the monotone class. A valid slot is obtained by placing the object to $s^\prime$ (line~19). If more than one object occlude the invalid slot, the objects except the last one which will be placed into the invalid slot (e.g., $o_0$ moved to $s_1$ in Fig.~\ref{fig:non_monotone_example}) will be placed to valid slots while following the principal using $\beta$ (lines~21--23).

Once a valid slot is obtained, the algorithm resumes to place objects in $\mathcal{O}_R$ to valid slots. After each rearrangement action, $\mathcal{S}_v$ and $\mathcal{O}_R$ are updated as they could change after every action (lines~6 and 31). At any time, the algorithm tries to obtain additional valid slots if they become insufficient. Example executions of Alg.~\ref{alg:rearrange} are shown in Fig.~\ref{fig:example_execution} for both monotone and non-monotone problems.

\begin{algorithm}
\caption{{\scshape RearrangementPlan}} \label{alg:rearrange}
\begin{algorithmic}[1]
{\small
\floatname{algorithm}{Procedure}
\renewcommand{\algorithmicrequire}{\textbf{Input:} }
\renewcommand{\algorithmicensure}{\textbf{Output: }}
\REQUIRE Workspace $\mathcal{W}$, objects $\mathcal{O}$, target $o_t$, candidate slots $\mathcal{S}_c$ 
\ENSURE 

\STATE $\mathcal{O}_R = $\textsc{RelocatePlan}$(\mathcal{W}, \mathcal{O}, o_t)$ \codecomment{find obstacles to be relocated (described in Sec.~\ref{sec:pre})}
\IF{$|\mathcal{O}_R| > |\mathcal{S}_c|$}
    \STATE \textbf{ return} Fail \codecomment{not enough empty space}
\ENDIF
\WHILE{$\mathcal{O}_R \neq \emptyset$}
    \STATE $\mathcal{S}_v$ = \textsc{FindValidCandidates}$(\mathcal{S}_c, \mathcal{W}, \mathcal{O}, \mathcal{O}_R, o_t)$ \codecomment{described in Sec.~\ref{sec:findvc}}
    \IF{$|\mathcal{S}_v| < |\mathcal{O}_R|$}
        \STATE Find all occluded candidate slots $\mathcal{S}_e$
        \IF{$\mathcal{S}_e = \emptyset$}
            \STATE \textbf{return} Fail
        \ENDIF
        \STATE $s^\prime = \argmin_{s_i \in \mathcal{S}_e} |\mathcal{O}^i_R|$\codecomment{Find $s_i$ that has the minimum number of objects to be relocated to access $s_i$}
        \STATE $\mathcal{O}^\prime_R = $\textsc{RelocatePlan}$(\mathcal{W}, \mathcal{O}, s^\prime)$
        \STATE $\mathcal{S}^\prime_v$ = \textsc{FindValidCandidates}$(\mathcal{S}_c, \mathcal{W}, \mathcal{O}, \mathcal{O}^\prime_R, s^\prime)$ 
        \IF{$|\mathcal{O}^\prime_R| -1 > |\mathcal{S}^\prime_v|$}
            \STATE \textbf{ return} Fail \codecomment{not enough empty slots to place objects in $\mathcal{O}^\prime_R$}
        \ENDIF
        
        \IF{$|\mathcal{O}^\prime_R| = 1 $}
            \STATE Move $o =$ \textsc{DeQueue}$(\mathcal{O}^\prime_R)$ to $s_i$
        \ELSE
            \STATE $\mathcal{B}^\prime$ = \textsc{ComputeBeta}$(\mathcal{S}^\prime_v, \mathcal{W}, \mathcal{O})$\codecomment{described in Sec.~\ref{sec:beta}}
            \STATE $(val, idx) = \min(\mathcal{B^\prime})$
            \STATE Move $o =$ \textsc{DeQueue}$(\mathcal{O}^\prime_R)$ to $s_{idx}$ 
        \ENDIF    
    \ELSE
        \STATE $\mathcal{B}$ = \textsc{ComputeBeta}$(\mathcal{S}_v, \mathcal{W}, \mathcal{O})$
        \STATE $(val, idx) = \min(\mathcal{B})$
        \STATE Move $o =$ \textsc{DeQueue}$(\mathcal{O}_R)$ to $s_{idx}$ 
    \ENDIF
    \STATE Update $\mathcal{O}$
    \STATE $\mathcal{O}_R = $\textsc{RelocatePlan}$(\mathcal{W}, \mathcal{O}, o_t)$
\ENDWHILE
\RETURN Success
}
\end{algorithmic}
\end{algorithm}

\smallskip
\thm \textbf{ 3.1} Task planning with Alg.~\ref{alg:rearrange} runs in polynomial time in the numbers of objects $N$ and candidate slots $M$ (without motion planning) for both of the monotone and non-monotone problems.

\pf Monotone: The modified VFH+ running in $O(N^2)$ is called $O(|\mathcal{S}_c|) = O(M)$ times to find valid candidates among all $|\mathcal{S}_c|$ candidates. Computing each $\beta(s_i)$ calls the modified VFH+ $|\mathcal{S}_v| \le M$ times to check accessibility of all valid candidates if $s_i$ is occupied. Finding $\beta$ for all valid candidates runs in $O(N^2 |\mathcal{S}_v|^2) = O(N^2 M^2)$. The above computation is repeated for every pick-and-place action which could be $N$ in the worst case (i.e., removing all objects). Therefore, the time complexity of Alg.~\ref{alg:rearrange} (excluding motion planning) is $O((N^2 M + N^2 M^2)N) = O(N^3 M^2)$.

Non-monotone: For the instances fall into the non-monotone class, valid candidates are acquired additionally by rearranging more objects. In the worst case, all candidate slots are invalid so there is no slot to place any of the  $\mathcal{O}_R$. Thus, the relocation algorithm~\cite{nam2019planning} can be used at most $|\mathcal{O}_R|$ times. Since the relocation algorithm runs in $O(N^4)$, this procedure takes $O(|\mathcal{O}_R|N^4) = O(N^5)$. Therefore, the time complexity of Alg.~\ref{alg:rearrange} (excluding motion planning) is $O((N^2 M + N^2 M^2 + N^5)N) = O(N^3 M^2 + N^6)$.
\qedm

Note that $M$ is not prohibitively large since the workspace is cluttered and confined ($M < 2N$ in all of our instances tested). If motion planning is included in the complexity analysis, the time complexity depends on the motion planner used (ranging from $O(n\log n)$ to $O(n^2)$ for both the process and query where $n$ is the number of samples). The motion planner is called for all candidate slots in the worst case to determine the validity of them. Thus, the motion planner is used at most $M$ times for each pick-and-place action.


\section{Experiments}
\label{sec:exp}

In this section, we show experimental results of our method. We first show how our method scales to the number of objects. Then we compare the number of pick-and-place actions, task and motion planning time, and the success rate with baseline methods using the Euclidean distance as the metric determining the slot to place objects. Lastly, we test our method in a simulated environment where dynamics of objects and the robot are modeled. The system is with Intel Core i7-6700 3.40GHz with 16G RAM and Python 2.7.

\subsection{Performance tests}

\begin{figure}[h!]
    \vspace{-5pt}
    \captionsetup{skip=0pt}
    \centering
    \includegraphics[scale=0.27]{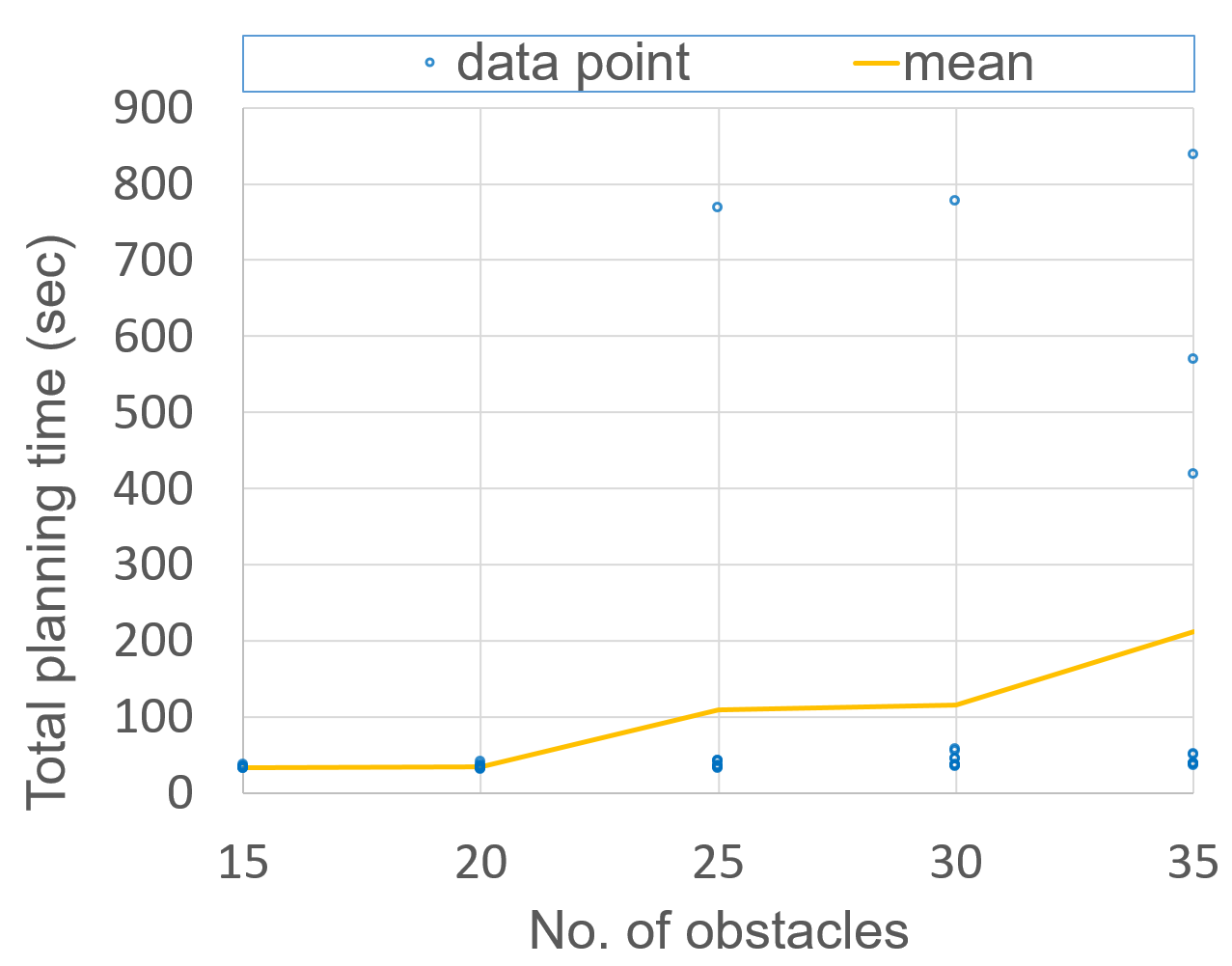}
    \caption{Offline task planning time without considering motion planning and execution}
    \label{fig:wo_mp_time}\vspace{-12pt}
\end{figure}

First, we test Alg.~\ref{alg:rearrange} offline without motion planning and execution so as to see how our task planning scales to the number of objects. Also, we define a large space ($l_w = 120cm$, and $l_d = 75cm$) to test instances with a large number of objects. The space with this size is not appropriate for real robots which has a fixed base owing to the limited workspace. 
The number of obstacles $N$ increases from 15 to 35 at intervals of 5. For each $N$, we run Alg.~\ref{alg:rearrange} with 10 random instances. If task and motion planning takes longer than five minutes, we terminate the algorithm and the instance is marked as failure. The results are shown in Fig.~\ref{fig:wo_mp_time}. In most cases, task planning is done within 100 seconds. Considering the time limit, the success rate when $N = 35$ is 75\%. In many practical instances, $N$ would not be larger than 25 where the algorithm shows 90\% of the success rate (the same success rate obtained when $N = 30$). The result shows that task planning scales to the number of obstacles in many real-world problems where $N$ is not prohibitively large. However, the test is done without motion planning so our method would require additional pick-and-place actions to deal with motion planning failures. 

Second, we test Alg.~\ref{alg:rearrange} with motion planning. Since there is no comparable method, we develop two baseline methods. Given a set of candidate slots, \textit{Farthest} method rearranges the obstacle to the farthest slot from the target and \textit{Deepest} method rearrange the obstacle to the farthest slot from the robot. Performance metrics are the success rate, the number of pick-and-place actions, and task and motion planning time. We use Kinova Jaco 1 and RRTConnect~\cite{kuffner2000rrt} from Open Motion Planning Library (OMPL)~\cite{sucan2012open} in MoveIt motion planning framework~\cite{moveit} for motion planning. The dimension of the workspace is $l_w = 90cm, l_d = 45cm, l_h = 45cm$.
We increase the number of obstacles from 9 to 15 at intervals of two. The time limit for task and motion planning is the same with the previous experiment (i.e., 5 mins). For each $N$, we first run the methods with 20 random instances and compute the success rate. Then we run additional instances to collect 20 data points for each method to have the equal sample size across all compared methods for statistics of the other performance metrics. 

\begin{figure}[h!]
    \vspace{-5pt}
    \captionsetup{skip=0pt}
    \centering
    \includegraphics[scale=0.26]{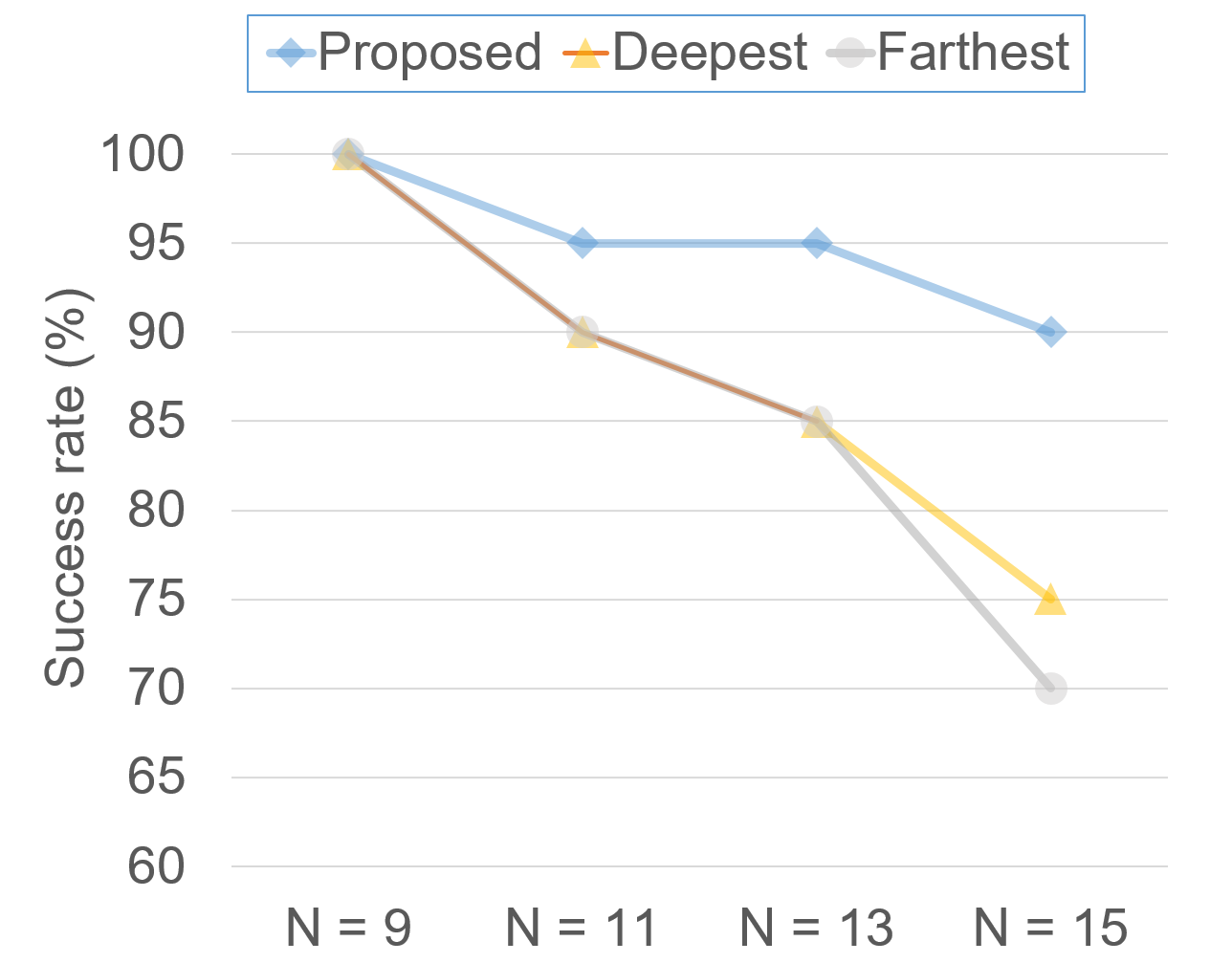}
    \caption{A comparison of the success rate. The proposed method maintains high success rates in difficult environments with a larger $N$.}
    \label{fig:mp_successrate}\vspace{-8pt}
\end{figure}

The success rates of the methods are shown in Fig.~\ref{fig:mp_successrate}. Our method shows higher success rates for $N \ge 11$. The difference in the success rate increases as $N$ increases. The largest difference occurs when $N=15$ which is 20\% compared to Farthest. Our method fails if the workspace does not have much free space to rearrange obstacles. Other methods does not only fail in the former case but also in all instances that fall into the non-monotone class. As $N$ increases, the environment becomes more complex and more non-monotone problems occur where the compared methods cannot solve.

\begin{figure}[h!]
    \captionsetup{skip=0pt}
    \centering
    \includegraphics[scale=0.26]{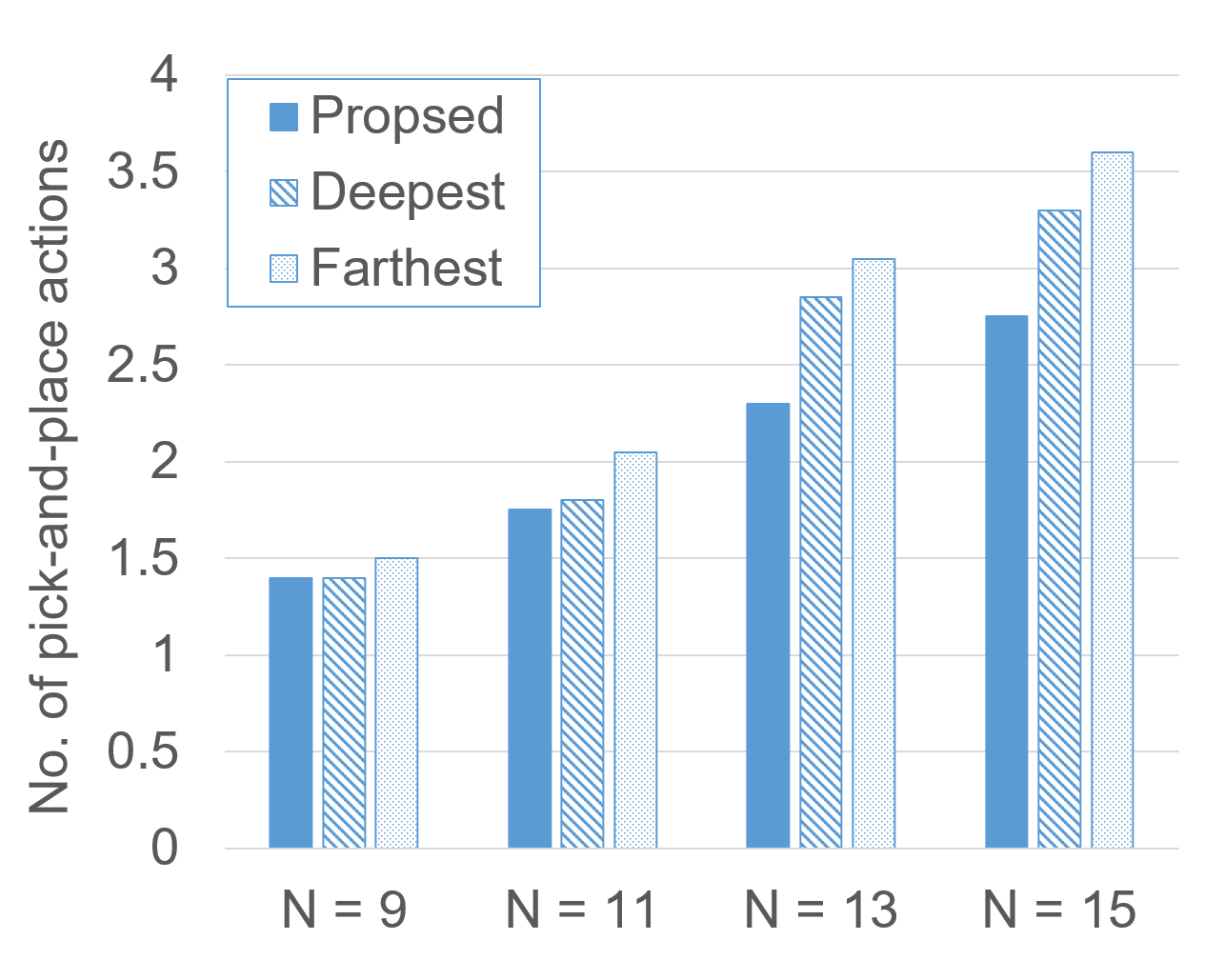}
    \caption{The number of pick-and-place actions. The proposed method shows the smallest value where the difference increases in more cluttered environments (as $N$ increases).}
    \label{fig:mp_action}\vspace{-12pt}
\end{figure}

\begin{figure}[h!]
    \captionsetup{skip=0pt}
    \centering
    \includegraphics[scale=0.26]{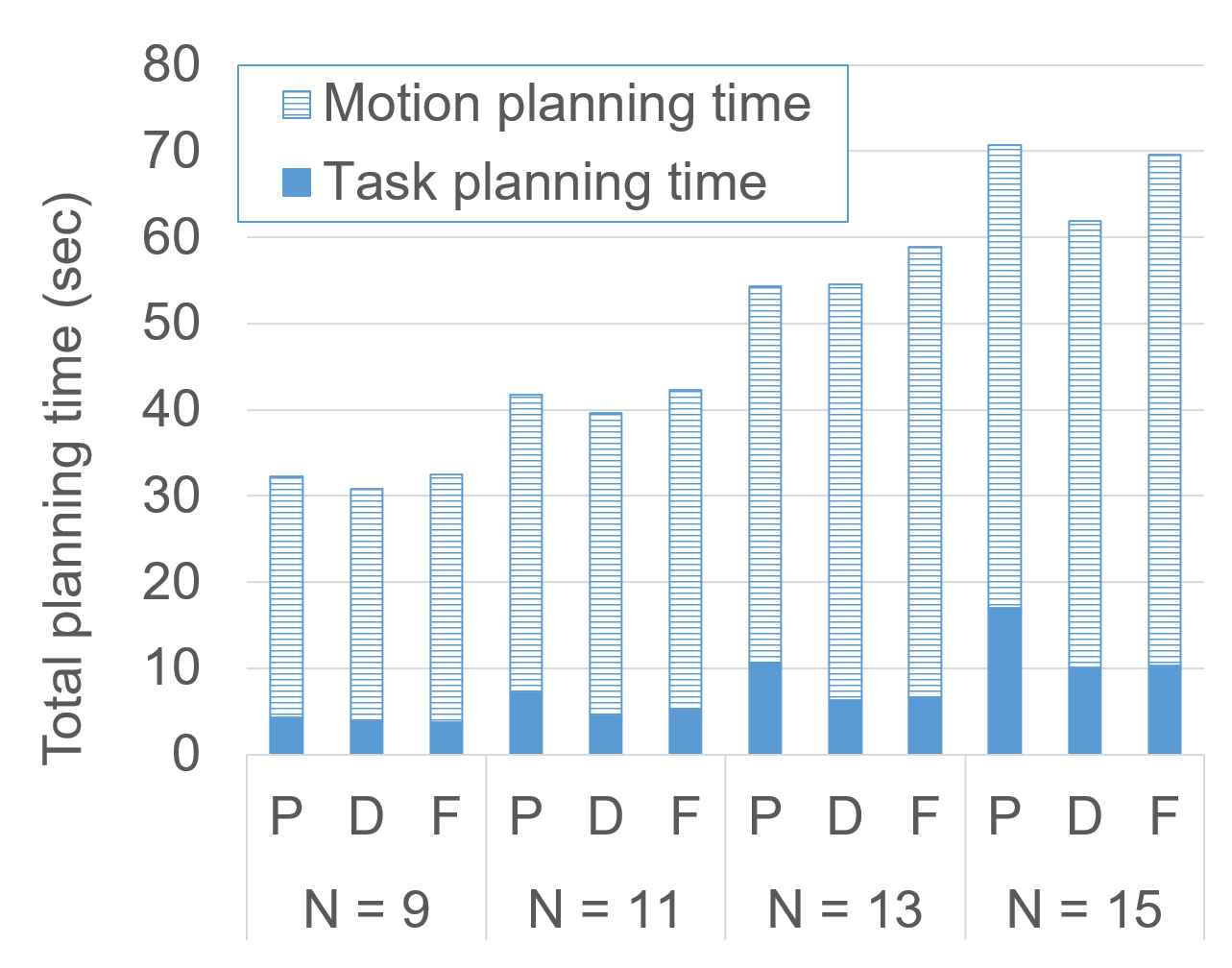}
    \caption{Task and motion planning time. Although the proposed method takes longer time in general, the differences with other methods are not statistically significant ($p < 0.05$ from 2-tailed t-tests).}
    \label{fig:mp_time}\vspace{-10pt}
\end{figure}

The performance metrics averaged from the collected 20 successful random instances are shown in Figs.~\ref{fig:mp_action}, \ref{fig:mp_time} and Table~\ref{tab:monotone result}. Note that the results only contain monotone instances as the compared method cannot solve non-monotone instances so comparison cannot be done with them. The results from non-monotone instances are shown in Table~\ref{tab:non-monotone result} where they are averaged from 2, 3, and 6 instances for $N=11, 13, 15$, respectively as non-monotone instances rarely occur with smaller $N$. Our method shows the smallest number of pick-and-place actions in average. The reductions of the number of actions are 16.6\% and 23.6\% when compared to Farthest and Deepest, respectively where $N=15$. A 2-tailed t-test shows that the difference between proposed method ($\mu = 2.75, \sigma = 1.16$) and Deepest ($\mu = 3.3, \sigma = 1.30$) is statistically significant ($p = 0.012$) when $N=15$. Farthest ($\mu = 3.6, \sigma = 1.78$) also shows a statistically significant difference ($p = 0.86$) with ours. Ours has smaller numbers of actions than others because they do not aim to utilize all candidate slots as many as possible while ours uses $\beta$ for choosing slots. Therefore, they sometimes perform actions that turns a monotone problem into non-monotone.

\begin{table*}[t!]
\centering
\captionsetup{skip=0pt}
\scalebox{0.9}{
\begin{tabular}{|c|c|c|c|c|c|c|c|c|c|c|c|c|c|}
\hline
\multicolumn{2}{|c|}{$N$}                                                           & \multicolumn{3}{c|}{9}                                                                                                                                                         & \multicolumn{3}{c|}{11}                                                                                                                                                        & \multicolumn{3}{c|}{13}                                                                                                                                                         & \multicolumn{3}{c|}{15}                                                                                                                                                        \\ \hline
\multicolumn{2}{|c|}{Method}                                                      & Proposed                                                     & Deep                                                     & Far                                                      & Proposed                                                     & Deep                                                     & Far                                                      & Proposed                                                     & Deep                                                     & Far                                                       & Proposed                                                     & Deep                                                     & Far                                                      \\ \hline
\multirow{3}{*}{\begin{tabular}[c]{@{}c@{}}Time \\ (sec)\end{tabular}}  & Overall & \begin{tabular}[c]{@{}c@{}}32.21 \\ (13.08)\end{tabular} & \begin{tabular}[c]{@{}c@{}}30.78 \\ (11.33)\end{tabular} & \begin{tabular}[c]{@{}c@{}}32.43 \\ (14.23)\end{tabular} & \begin{tabular}[c]{@{}c@{}}41.71 \\ (20.05)\end{tabular} & \begin{tabular}[c]{@{}c@{}}39.57 \\ (14.80)\end{tabular} & \begin{tabular}[c]{@{}c@{}}42.26 \\ (16.73)\end{tabular} & \begin{tabular}[c]{@{}c@{}}54.37 \\ (14.04)\end{tabular} & \begin{tabular}[c]{@{}c@{}}54.54 \\ (18.81)\end{tabular} & \begin{tabular}[c]{@{}c@{}}54.84 \\ (19.60)\end{tabular}  & \begin{tabular}[c]{@{}c@{}}70.65 \\ (35.94)\end{tabular} & \begin{tabular}[c]{@{}c@{}}61.87 \\ (24.82)\end{tabular} & \begin{tabular}[c]{@{}c@{}}69.62 \\ (30.37)\end{tabular} \\ \cline{2-14} 
                                                                        & Task    & \begin{tabular}[c]{@{}c@{}}4.33 \\ (2.56)\end{tabular}   & \begin{tabular}[c]{@{}c@{}}3.87 \\ (0.96)\end{tabular}   & \begin{tabular}[c]{@{}c@{}}3.70 \\ (1.12)\end{tabular}   & \begin{tabular}[c]{@{}c@{}}7.15 \\ (4.61)\end{tabular}   & \begin{tabular}[c]{@{}c@{}}4.54 \\ (1.30)\end{tabular}   & \begin{tabular}[c]{@{}c@{}}5.19 \\ (1.44)\end{tabular}   & \begin{tabular}[c]{@{}c@{}}10.65 \\ (4.20)\end{tabular}  & \begin{tabular}[c]{@{}c@{}}6.33 \\ (1.75)\end{tabular}   & \begin{tabular}[c]{@{}c@{}}6.58 \\ (1.93)\end{tabular}    & \begin{tabular}[c]{@{}c@{}}17.02 \\ (10.31)\end{tabular} & \begin{tabular}[c]{@{}c@{}}10.04 \\ (4.24)\end{tabular}  & \begin{tabular}[c]{@{}c@{}}10.30 \\ (4.57)\end{tabular}  \\ \cline{2-14} 
                                                                        & Motion  & \begin{tabular}[c]{@{}c@{}}27.88 \\ (10.86)\end{tabular} & \begin{tabular}[c]{@{}c@{}}26.90 \\ (10.53)\end{tabular} & \begin{tabular}[c]{@{}c@{}}28.72 \\ (13.23)\end{tabular} & \begin{tabular}[c]{@{}c@{}}34.55 \\ (16.22)\end{tabular} & \begin{tabular}[c]{@{}c@{}}35.02 \\ (13.70)\end{tabular} & \begin{tabular}[c]{@{}c@{}}37.07 \\ (15.77)\end{tabular} & \begin{tabular}[c]{@{}c@{}}43.72 \\ (10.47)\end{tabular} & \begin{tabular}[c]{@{}c@{}}48.20 \\ (17.20)\end{tabular} & \begin{tabular}[c]{@{}c@{}}52.261 \\ (17.82)\end{tabular} & \begin{tabular}[c]{@{}c@{}}53.63 \\ (25.95)\end{tabular} & \begin{tabular}[c]{@{}c@{}}51.83 \\ (20.96)\end{tabular} & \begin{tabular}[c]{@{}c@{}}59.32\\ (25.88)\end{tabular}  \\ \hline
\multicolumn{2}{|c|}{\begin{tabular}[c]{@{}c@{}}Number of\\ actions\end{tabular}} & \begin{tabular}[c]{@{}c@{}}1.40 \\ (0.59)\end{tabular}   & \begin{tabular}[c]{@{}c@{}}1.40 \\ (0.59)\end{tabular}   & \begin{tabular}[c]{@{}c@{}}1.50 \\ (0.76)\end{tabular}   & \begin{tabular}[c]{@{}c@{}}1.75 \\ (0.63)\end{tabular}   & \begin{tabular}[c]{@{}c@{}}1.8 \\ (0.76)\end{tabular}    & \begin{tabular}[c]{@{}c@{}}2.05 \\ (0.99)\end{tabular}   & \begin{tabular}[c]{@{}c@{}}2.3 \\ (0.57)\end{tabular}    & \begin{tabular}[c]{@{}c@{}}2.85 \\ (1.18)\end{tabular}   & \begin{tabular}[c]{@{}c@{}}3.05 \\ (1.14)\end{tabular}    & \begin{tabular}[c]{@{}c@{}}2.75 \\ (1.16)\end{tabular}   & \begin{tabular}[c]{@{}c@{}}3.3 \\ (1.30)\end{tabular}    & \begin{tabular}[c]{@{}c@{}}3.6\\ (1.78)\end{tabular}     \\ \hline
\end{tabular}
}
\caption{The statistics from 20 monotone instances showing the mean and standard deviation (in parentheses). All the methods succeed in all instances so the success rate is omitted.}
\label{tab:monotone result}
\vspace{-10pt}
\end{table*}

\begin{table}[h!]
\centering
\captionsetup{skip=0pt}
\scalebox{0.87}{\begin{tabular}{|c|c|c|c|c|}
\hline
\multicolumn{2}{|c|}{$N$}                                                           & 11                                                       & 13                                                        & 15                                                        \\ \hline
\multirow{3}{*}{\begin{tabular}[c]{@{}c@{}}Time \\ (sec)\end{tabular}}  & Overall & \begin{tabular}[c]{@{}c@{}}83.62 \\ (17.57)\end{tabular} & \begin{tabular}[c]{@{}c@{}}142.66 \\ (40.62)\end{tabular} & \begin{tabular}[c]{@{}c@{}}167.45 \\ (43.01)\end{tabular} \\ \cline{2-5} 
                                                                        & Task    & \begin{tabular}[c]{@{}c@{}}19.61 \\ (1.85)\end{tabular}  & \begin{tabular}[c]{@{}c@{}}29.24\\ (9.09)\end{tabular}    & \begin{tabular}[c]{@{}c@{}}49.41 \\ (16.75)\end{tabular}  \\ \cline{2-5} 
                                                                        & Motion  & \begin{tabular}[c]{@{}c@{}}64.01 \\ (15.71)\end{tabular} & \begin{tabular}[c]{@{}c@{}}113.42 \\ (31.98)\end{tabular} & \begin{tabular}[c]{@{}c@{}}118.04 \\ (28.29)\end{tabular} \\ \hline
\multicolumn{2}{|c|}{\begin{tabular}[c]{@{}c@{}}Number of\\ actions\end{tabular}} & \begin{tabular}[c]{@{}c@{}}3.5 \\ (0.707)\end{tabular}   & \begin{tabular}[c]{@{}c@{}}6.0\\ (1.73)\end{tabular}      & \begin{tabular}[c]{@{}c@{}}8.66 \\ (1.16)\end{tabular}    \\ \hline
\end{tabular}
}
\caption{The results from non-monotone instances. Other methods are not compared as they all fail with non-monotone instances. There are 2, 3, and 6 samples for $N = 11, 13, 15$, respectively.}
\label{tab:non-monotone result}
\vspace{-10pt}
\end{table}

Ours shows longer task and motion planning time compared to the other methods (e.g., 8.78 sec and 1.03 sec longer than Farthest and Deepest, respectively when $N=15$). However a 2-tailed t-test shows that the planning times of ours ($\mu = 70.65, \sigma = 35.94$) and Deepest ($\mu = 61.87, \sigma = 24.82$) do not have a difference which is statistically significant ($p = 0.1081$). Similarly, ours planning time is not different from Farthest ($\mu = 69.62, \sigma = 30.37$) significantly ($p = 0.86$).

Table~\ref{tab:non-monotone result} shows the result of the proposed method solving non-monotone instances. Our proposed method fails for two reasons in the non-monotone problem. First, it fails when the workspace has insufficient candidate slots. The failing cases could be reduced if more candidate slots can be found before our method begins. Second, an excessive number of trials for motion planning in clutter could increase planning time beyond the time limit. This type of failure can be reduced if motion planning success rates increase with practical schemes like trying with different poses of the end-effector if motion planning fails.



\subsection{Experiments in a simulated environment}
\label{sec:sim}

\begin{figure}[h!]
    \centering
    \captionsetup{skip=0pt}
    \includegraphics[scale=0.30]{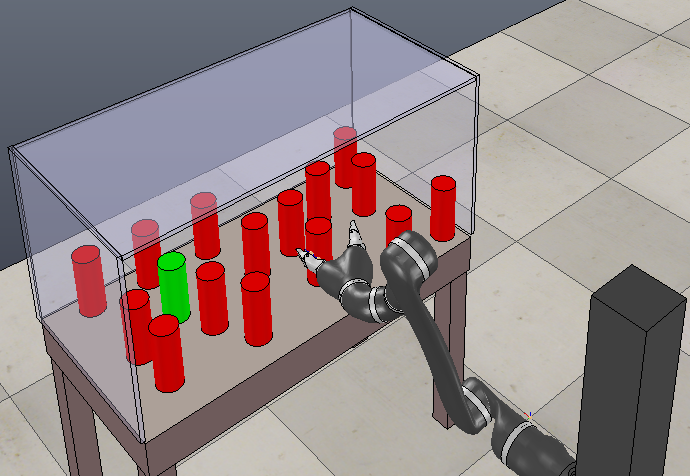}
    \caption{The environment for a realistic simulation. We use a model of Kinova Jaco1 in the simulator V-REP. We test with 15 red obstacles and one green target object.}
    \label{fig:vrep_env}\vspace{-10pt}
\end{figure}

Through realistic simulations, we want to show how our method could work in real world and how much the total execution time can be reduced. We use a high-fidelity simulator V-REP~\cite{rohmer2013v} with Vortex Dynamics physics engine. We use a model of Kinova Jaco1, a 6-DOF manipulator anchored at a base station (Fig.~\ref{fig:vrep_env}).  
We run Alg.~\ref{alg:rearrange} with 10 random instances for $N=15$.
Our method shows the smallest number of pick-and-place actions and execution time in average. The reductions of the number of actions are 13.7\% and 23.1\% and the execution time are 19.6\% and 28.1\% when compared to Deepest and Farthest, respectively. 
The execution time in the proposed method ($\mu = 301.53, \sigma = 97.72$) and Deepest ($\mu = 375.20, \sigma = 186.07$) are different with statistical significance ($p = 0.047$). Farthest ($\mu = 410.2, \sigma = 186.2$) also shows a statistically significant difference ($p = 0.02$) with ours. The experiment shows that our method could finish the rearrangement problem efficiently in real world applications. 


\begin{table}[h!]
\vspace{-5pt}
\centering
\captionsetup{skip=0pt}
\scalebox{0.87}{
\begin{tabular}{|c|c|c|c|}
\cline{1-4}
$N$ & \multicolumn{3}{c|}{15} \\
\hline
Method & Proposed & Deep  & Far \\
\hline
{\begin{tabular}[c]{@{}c@{}}Time\\ (sec)\end{tabular}}        & {\begin{tabular}[c]{@{}c@{}}301.53 \\ (97.72)\end{tabular}} & {\begin{tabular}[c]{@{}c@{}}375.20 \\ (186.07)\end{tabular}} &  {\begin{tabular}[c]{@{}c@{}}410.28 \\ (186.29)\end{tabular}} \\ 
\hline
{\begin{tabular}[c]{@{}c@{}}Number of\\ actions\end{tabular}} & {\begin{tabular}[c]{@{}c@{}}2.5 \\ (0.707)\end{tabular}}    & {\begin{tabular}[c]{@{}c@{}}2.9 \\ (0.99)\end{tabular}}      & {\begin{tabular}[c]{@{}c@{}}3.2 \\ (0.91)\end{tabular}}    \\
\hline
\end{tabular}
}
\caption{The result from simulations using a dynamic simulator. The statistics are from 10 monotone instances. The numbers are the means and standard deviations (in parentheses).}
\label{tab:vrep_test}
\vspace{-10pt}
\end{table}

\section{Conclusion}
In this work, we solve the rearrangement problem of objects in cluttered and confined spaces.  Especially, we consider the problem of \textit{where to place} objects that should be relocated to secure a collision-free path to the target object. Our task planning for rearrangement is combined with motion planning considering the kinematic constraints of the whole robot arm. Our method follows a principle to choose the slot to place objects in which the placement maximally utilizes empty spaces in the workspace. Our method can solve the monotone problem as well as the non-monotone problem which is more difficult. Experimental results show that ours reduces the number of pick-and-place actions and the total execution time for object rearrangement. Our future work includes obtaining more candidate slots using non-prehensile actions or stacking objects. Also, we will consider different shapes of objects where grasping is allowed at some particular directions so rearrangement becomes harder.

\bibliographystyle{IEEEtran}
\bibliography{references}
\end{document}